\documentclass[10pt,twocolumn,]{article}

\usepackage{cvpr}
\usepackage{times}
\usepackage{epsfig}
\usepackage{graphicx}
\usepackage{amsmath}
\usepackage{bm}
\usepackage{amssymb}
\usepackage{amsfonts}
\usepackage{booktabs}
\usepackage{caption}
\usepackage{subcaption}
\usepackage{physics}
\usepackage{bbm}
\usepackage{etoolbox}
\robustify\bfseries

\usepackage{footnote}
\makesavenoteenv{tabular}

\newenvironment{tight_itemize}{
\begin{itemize}
  \setlength{\itemsep}{0pt}
  \setlength{\parskip}{0pt}
  \setlength{\topsep}{0pt}
  \setlength{\partopsep}{0pt}
}{\end{itemize}}

\usepackage[pagebackref=true,breaklinks=true,letterpaper=true,colorlinks,bookmarks=false]{hyperref}

\cvprfinalcopy 

\def\cvprPaperID{2698} 

\ifcvprfinal\fi
\begin{document}

\title{Parameter Reference Loss for Unsupervised Domain Adaptation}

\author{Jiren Jin$^1$\thanks{Part of this work was done when Jiren Jin was an intern at Preferred Networks, Inc.}, Richard G\@.~Calland$^2$, Takeru Miyato$^2$, Brian K\@.~Vogel$^2$, Hideki Nakayama$^1$\\
$^1$The University of Tokyo \quad $^2$Preferred Networks, Inc. \\
{\tt\small \{jin,nakayama\}@nlab.ci.i.u-tokyo.ac.jp} \quad {\tt\small \{calland,miyato,vogel\}@preferred.jp}}

%


\maketitle
\thispagestyle{empty}
\begin{abstract}
The success of deep learning in computer vision is mainly attributed to an abundance of data. However, collecting large-scale data is not always possible, especially for the supervised labels. Unsupervised domain adaptation (UDA) aims to utilize labeled data from a source domain to learn a model that generalizes to a target domain of unlabeled data. 
A large amount of existing work uses Siamese network-based models, where two streams of neural networks process the source and the target domain data respectively. Nevertheless, most of these approaches focus on minimizing the domain discrepancy, overlooking the importance of preserving the discriminative ability for target domain features.
Another important problem in UDA research is how to evaluate the methods properly. Common evaluation procedures require target domain labels for hyper-parameter tuning and model selection, contradicting the definition of the UDA task. Hence we propose a more reasonable evaluation principle that avoids this contradiction by simply adopting the latest snapshot of a model for evaluation. This adds an extra requirement for UDA methods besides the main performance criteria: the stability during training.
We design a novel method that connects the target domain stream to the source domain stream with a Parameter Reference Loss (PRL) to solve these problems simultaneously. Experiments on various datasets show that the proposed PRL not only improves the performance on the target domain, but also stabilizes the training procedure. As a result, PRL based models do not need the contradictory model selection, and thus are more suitable for practical applications.
\end{abstract}

\section{Introduction}

\begin{figure}[t]
\begin{center}
   \includegraphics[width=\linewidth]{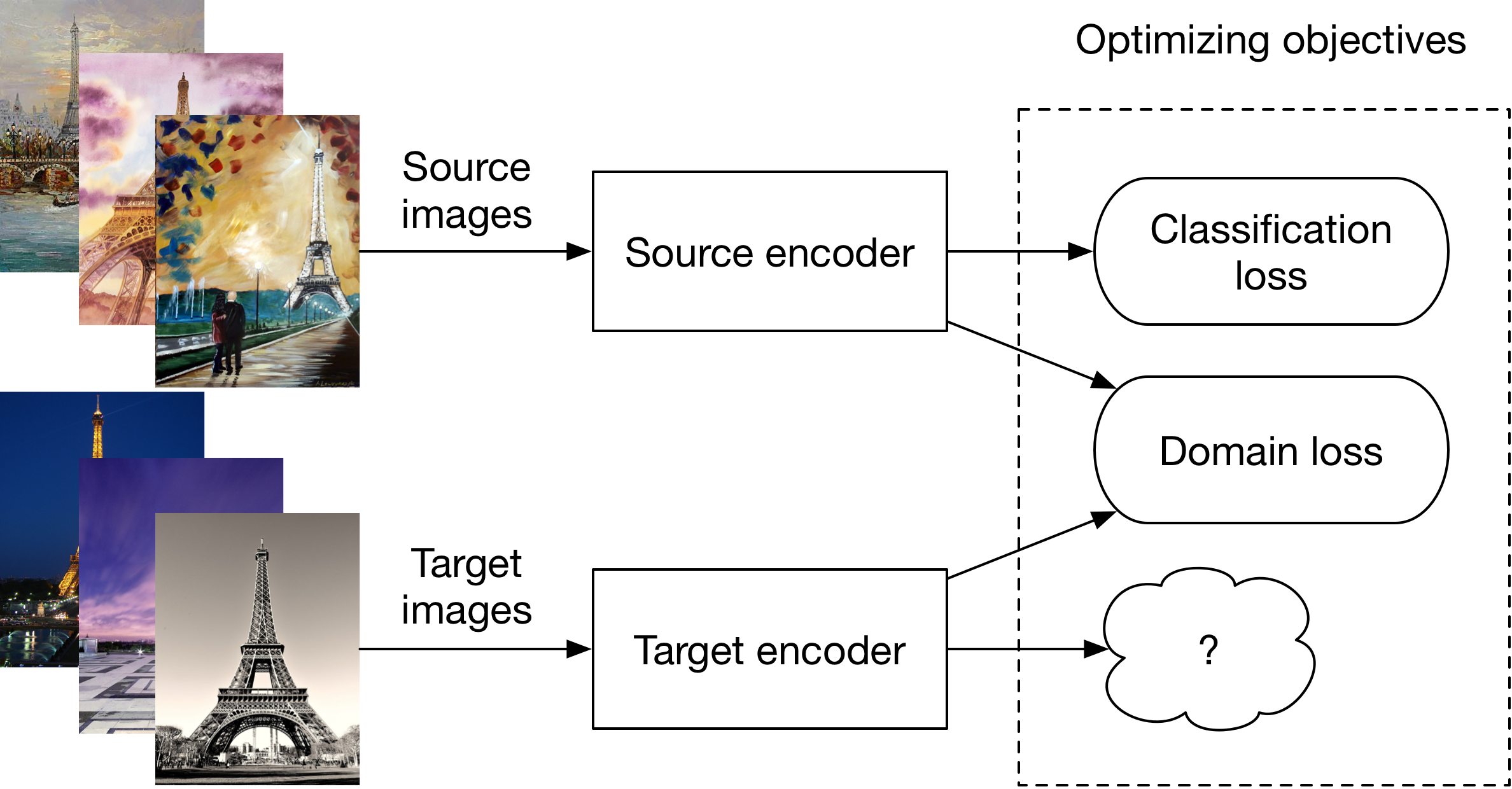}
\end{center}
\vspace{-0.4cm}
\caption{The general architecture of Siamese network based unsupervised domain adaptation models. The classification loss is only applied to the source domain encoder since no labels are available on the target domain. The domain loss can be any loss function that is able to minimize the discrepancy between two domains, including common discrepancy loss and the recent popular adversarial loss. The question mark indicates the missing component of existing work: lacking constraints to preserve the discriminative ability for target domain features.}
\label{fig:siamese}
\vspace{-0.5cm}
\end{figure}

The availability of large-scale data is known to be one of the critical success factors of deep learning~\cite{lecun2015deep}. As shown by Sun \etal~\cite{sun2017revisiting}, increasing the amount of training data almost always improves the performance of a deep model. Nevertheless, collecting large data sets for many specific real world applications is still difficult and expensive due to the labor-intensive workload. Particularly, labeled data is more difficult to collect than raw data, like the ground-truth categories for image classification. 

Domain adaptation~\cite{csurka2017domain}, tries to take advantage of available labeled data from a source domain to learn a model on a target domain where few (or no) labels are available.
Domain adaptation is needed because the assumption of identically and independently distributed (i.i.d) data is usually not satisfied in real world applications, \ie, data in the target/deploying phase is drawn from a distribution different from that of the source training data. In this case the dataset bias is usually caused by the data collection procedure~\cite{torralba2011unbiased}. Sometimes we might want to intentionally train the model on a different domain to help improve generalization on the target domain, \eg, training a model with synthetic data to improve the performance on real world data~\cite{bousmalis2017using}.

In this work we focus on the case where no label information is available for the target domain, which is often referred to as Unsupervised Domain Adaptation (UDA). There has already been much progress~\cite{gopalan2011domain, pan2011domain, gong2012geodesic, long2013transfer, fernando2013unsupervised, sun2016return} before the use of deep models. Recent trends involve combining traditional algorithms with deep features~\cite{donahue2014decaf}, as well as designing novel architectures for deep domain adaptation~\cite{chopra2013dlid, tzeng2014deep, long2015learning, long2016deep, ganin2016domain, tzeng2017adversarial, bousmalis2016unsupervised, ghifary2016deep, bousmalis2016domain}. Siamese networks~\cite{bromley1994signature} are the most commonly used basic architecture for these methods, where two encoders are used for the source and target domain data~\cite{tzeng2017adversarial} respectively, as shown in Figure~\ref{fig:siamese}.

\subsection{Problems and our solution}
Most existing methods focus on minimizing the domain discrepancy, but overlook the importance of preserving the discriminative ability for the target domain features. Note the differences between the constraints for the source domain encoder and target domain encoder in Figure~\ref{fig:siamese}: two constraints for the source domain encoder, a label classification loss, and a domain loss; while only one constraint for the target domain encoder, the domain loss. The missing classification loss constraint may cause the learned target domain features to lack discriminative ability, since these features are only optimized to match the source domain distribution. Such insufficiency comes from the lack of labels on the target domain, which prevents a direct connection between the target domain features and the label classifier.

A natural solution would be to add some additional constraints to the target domain encoder so as to encourage it to preserve important information for discriminating between different classes. Due to the unsupervised nature of the target domain, a reconstruction loss first comes to mind. However, unlike the common unsupervised learning or the recent style transfer tasks~\cite{gatys2015neural, zhu2017unpaired}, preserving pixel-to-pixel information contradicts the objective of learning domain invariant features. On the other hand, directly aligning the features from both domains is also not applicable, since the two domain inputs are not paired, \ie, no explicit matching between the features of two domains exists. 

To learn a more meaningful representation for the target domain data, we propose a novel Parameter Reference Loss (PRL) to build a flexible connection between the source domain encoder and the target domain encoder. Furthermore, we show that PRL can improve the training stability, which solves another important problem: the contradiction of using target domain labels for model selection in UDA. A detailed discussion of why this problem matters and how PRL helps can be found in Section~\ref{stable_uda}.

Another motivation for PRL is that we think the current use of learned parameters wastes resources because these parameters are often only used for initializing another model for a new domain/task. Hence we try to make the model able to benefit from the previous learned parameters even during the adaptation training phase. In fact, more efficiently using such resources plays a more important role especially for the UDA task, as a result of the absence of target domain labels.

\subsection{Contributions}
In summary, the contributions of this work include:
\begin{tight_itemize}
	\item We point out the problem of poor discriminative ability caused by the lack of constraint for the target domain encoder in existing work.
	\item We clarify the contradiction of evaluation procedures for UDA methods, and propose a direction to solve this problem: stabilize the training.

	\item We propose a solution to solve the above problems simultaneously using PRL, which can be easily combined with most existing UDA methods that are based on Siamese networks.

	\item We show that previously learned parameters can be more useful during the training phase than simply using them for model initialization. \end{tight_itemize}

\section{Related work}
We review recent deep learning based domain adaptation methods since they are most related to our proposed method.

As the main objective of domain adaptation methods is to learn a representation that is invariant to domain change, we can categorize existing methods into two groups according to the loss function used for minimizing the domain discrepancy.

\subsection{Discrepancy loss based domain adaptation}
The first group of methods uses discrepancy loss like Maximum Mean Discrepancy (MMD)~\cite{gretton2012kernel} to learn domain invariant features. Tzeng \etal proposed Deep Domain Confusion~\cite{tzeng2014deep}, one of the first domain adaptation methods based on deep neural networks. They apply the AlexNet~\cite{krizhevsky2012imagenet} model to both source and target domain inputs, explicitly minimizing the discrepancy loss between the extracted features using MMD. Deep Adaptation Networks (DAN)~\cite{long2015learning} extends this work using multi-kernel MMD on three different layers, arguing that minimizing the discrepancy on the last layer is not sufficient to remove the domain difference caused by the early layers.

Besides MMD loss, Deep CORAL~\cite{sun2016deep} minimizes the domain discrepancy by aligning the correlations of activation layers in the deep model. Zellinger~\etal~\cite{zellinger2017central} propose Central Moment Discrepancy (CMD), an explicit order-wise matching of higher order moments, to avoid computationally expensive distance and kernel matrix computations.
Csurka~\etal~\cite{csurka2017discrepancy} have done a comparative study on the discrepancy based UDA models using various deep features.

\begin{figure*}[t]
\begin{center}
   \includegraphics[width=\linewidth]{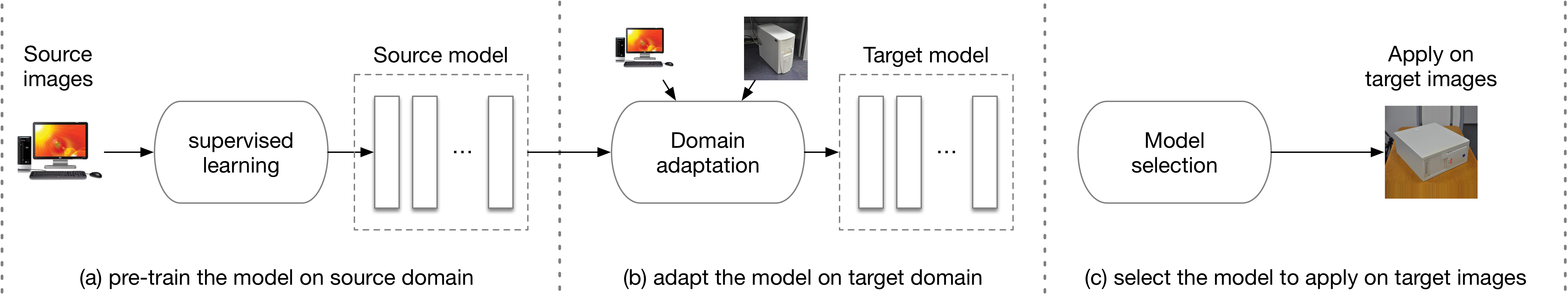}
\end{center}
   \caption{A typical workflow for unsupervised domain adaptation on an image classification task. First the model is trained only on the source domain images with labels using a supervised learning approach, where we can use common cross-validation techniques for hyper-parameter tuning and model selection. The second step is to adapt the trained model on the target domain images using domain adaptation methods. Note that though target labels are not available for use by definition of the UDA task, it has been common practice that the target domain labels are used for both evaluation and model selection. However, we argue that the comparison of different UDA methods based on model selection using the target domain labels does not accurately reflect the performance of the evaluated methods.}
\label{fig:workflow}
\vspace{-0.2cm}
\end{figure*}

\subsection{Adversarial loss based domain adaptation}
The adversarial loss has been recently popularized by Generative Adversarial Networks (GANs)~\cite{goodfellow2014generative}. 
Bousmalis \etal~\cite{bousmalis2016unsupervised} propose to use GANs to generate target domain data conditioned on the source domain inputs. 
Russo \etal~\cite{russo2017source} adopt CycleGAN~\cite{zhu2017unpaired} for domain adaptation to tackle the problem caused by the unpaired inputs of both domains. 

The adversarial loss can also be combined with discriminative models. ReverseGrad~\cite{ganin2015unsupervised} applies the adversarial loss on the features extracted by a discriminative model. The implementation of ReverseGrad uses a gradient reverse layer to compute the gradients for the encoder, which is indeed a different way to compute the adversarial loss for the generator (encoder) part.  

\subsection{General UDA model}

Tzeng \etal~\cite{tzeng2017adversarial} summarize the methods using Siamese networks and adversarial loss. They categorize these methods according to three design options: 1) whether the parameters are shared or not in the Siamese architecture, 2) whether the models are discriminative or generative, and 3) the choice of the adversarial loss. Adding discrepancy loss to the third option leads the above summarization to a general architecture for UDA tasks. 

For methods using a single encoder (generator) for both domains, the parameter sharing mechanism helps preserve the discriminative ability learned from the source domain data, however, it also limits the flexibility of the target domain model. For methods using independent encoders, preserving the discriminative ability of target domain features is often overlooked. In the specific model designed in ADDA~\cite{tzeng2017adversarial}, the target domain encoder is initialized using the parameters from the source domain encoder. This alleviates the problem caused by the single domain related constraint on the target encoder, but it is not enough to maintain the discriminative ability for the target domain features as the training continues.


\section{Stabilizing UDA}
\label{stable_uda}
We first give a formal definition of unsupervised domain adaptation for the image classification task to facilitate the explanation of our proposed method. Most of the definition is borrowed from Pan \etal~\cite{pan2010survey}. 

A domain $\mathcal{D}$ consists of two components: a feature space $\mathcal{X}$ with $m$-dimensionality and a marginal distribution $P(X)$, where $X = \{x_1,\dots,x_n\} \in \mathcal{X}$. Specifically, the feature spaces are the same in our task (image pixels or extracted features), thus the differences between domains are caused by different marginal probability distributions.

Given a specific domain, $\mathcal{D} = \{\mathcal{X}, P(X)\}$, a task $\mathcal{T}$ consists of two components: a label space $\mathcal{Y}$ with $K$-cardinality and an objective predictive function $f(\cdot)$. From a probabilistic viewpoint, $f(x)$ can also be written as $P(y|x)$.

We consider two different domains, the source domain $\mathcal{D}_S$ and the target domain $\mathcal{D}_T$. In domain adaptation, the label space is generally assumed to be the same for both domains, \ie, $\mathcal{Y}_S = \mathcal{Y}_D$. In the image classification task, this means the possible classes for each domain are the same, and hence we use $\mathcal{Y}$ to denote the label space of both domains. The source domain dataset is denoted as $D_S = \{ (x_{S_1}, y_{1}), \dots, (x_{S_{n_S}}, y_{S_{n_S}}) \}$, where $x_{S_i} \in \mathcal{X}_S$ is the image instance and $y_{S_{i}} \in \mathcal{Y}$ is the corresponding class label for that image. The target domain dataset is denoted in a similar way, with a key difference that the labels are not available: $D_T = \{ x_{T_1}, \dots, x_{T_{n_T}} \}$. 

The objective of unsupervised domain adaptation is to learn a model that can predict the labels for the target domain data by utilizing the source domain data and labels, with only the target domain data. In particular for the image classification task, the objective is to correctly predict the category of the given target domain image, \ie, $P(y|x_T)$.

\subsection{Contradiction of evaluating UDA methods}
Figure~\ref{fig:workflow} illustrates the typical workflow for unsupervised domain adaptation on an image classification task. The first step usually involves training the model on the source domain dataset only. The second step involves adapting the trained model to the target domain. Some methods combine these two steps to simultaneously learn for classification and adaptation. Note that the difficulties of unsupervised domain adaptation tasks include not only the unavailability of directly training the model with supervised information on the target domain, but also the contradiction of the hyper-parameter tuning and model selection procedure.

By the definition of unsupervised domain adaptation, it is impossible to use the target domain labels for validation purposes or selecting hyper-parameters. The simplest solution to avoid such a contradiction is just not to do hyper-parameter tuning and model selection. However, UDA methods are generally more sensitive to hyper-parameter changes compared to supervised learning approaches. As a result, besides using the complicated reverse cross validation~\cite{zhong2010cross}, the only feasible option to obtain reliable performance would be to use labeled supervision on the target domain for hyper-parameter optimization, as far as we know. Nevertheless, using target domain labels for hyper-parameter tuning biases the reported accuracy and does not accurately reflect the performance in real world tasks, where the stability of models might be more important than the possible performance gain. 

As completely avoiding hyper-parameter tuning and model selection is difficult, we consider from another direction to make the hyper-parameter tuning and model selection procedure easier. We will show that we can avoid the above contradiction if we can tune the hyper-parameters without looking at the target domain labels, and stabilize the adaptation training procedure so that no large performance drop is expected. Hence we modify the existing evaluation procedure to avoid the contradictions of using target domain labels:

\begin{itemize}
	\item During the adaptation training phase, select hyper-parameters without access to the target domain labels.

	\item Given a fixed number of training epochs, always select the latest epoch/snapshot of the trained model for the final evaluation or deployment.
\end{itemize}

To fulfill the above requirements, as well as to overcome the previous problem of lack of discriminative power, we propose the Parameter Reference Loss, which we explain in detail in the next Section.

\section{Parameter reference loss}
We first describe a baseline model to realize a typical unsupervised domain adaptation method using deep neural networks. Then we explain the proposed Parameter Reference Loss and its variants in detail.

\subsection{Baseline model}
The baseline method we used has a similar architecture to that of ADDA~\cite{tzeng2017adversarial}, where a Neural Network model $E_S$ is first trained on the source domain for classification using cross-entropy loss, and then the parameters of this model are used for initializing the target domain encoder $E_T$ having the same architecture. During the adaptation process, the source domain encoder $E_S$ and the classifier $C$ remain fixed while the target domain encoder is trained to produce features that are similar to the source domain features using the adversarial loss proposed in \cite{goodfellow2014generative}.

The reason that we use a different discrepancy loss instead of directly using exactly the same model in ADDA with adversarial loss is related to the ``more reasonable'' evaluation setting. Compared with adversarial loss, the discrepancy loss is less sensitive to be used as a metric for unsupervised hyper-parameter tuning when other hyper-parameters are the same. The reason that adversarial loss is not sufficient to measure the current domain discrepancy is because there are two loss terms for the generator and discriminator respectively, which influence each other. Moreover, training with the adversarial loss has issues of instability due to the complex mini-max optimization, and currently still needs much effort to tune the model to work well~\cite{salimans2016improved}.

There are usually two loss terms for the source domain model, a label classification loss and a discrepancy loss, however, since the source domain encoder of the baseline model is fixed during the adaptation phase, it is actually only being optimized with the classification loss when pre-training on the source domain data. Some other models described in Section~\ref{experiment} do have the two loss terms for the source domain encoder during adaptation.
The classification loss is defined as the cross-entropy loss:
\begin{multline}
	\mathcal{L}_{CLS}(X_S, Y_S, E_S) = \\
	- \mathbb{E}_{(x_S, y_S)\sim(X_S, Y_S)} \sum_{k=1}^{K}{\mathbbmss{1}_{[k=y_S]} \log C(E_S(x_S))}.
\end{multline}%

And the domain discrepancy loss is defined as the Maximum Mean Discrepancy (MMD) loss:
\begin{multline}
	\mathcal{L}_{MMD}(X_S, X_T, E_S, E_T) = \\
	\norm{\frac{1}{n_S}\sum_{i=1}^{n_S}{\phi(E_S(x_{S_i}))} - 
	\frac{1}{n_T}\sum_{j=1}^{n_T}{\phi(E_T(x_{T_j}))} }_\mathcal{H},
\end{multline}%
where $\phi(\cdot)$ is the RKHS kernel~\cite{gretton2012kernel}.

\subsection{Naive PRL}

\begin{figure}[t]
\begin{center}
   \includegraphics[width=\linewidth]{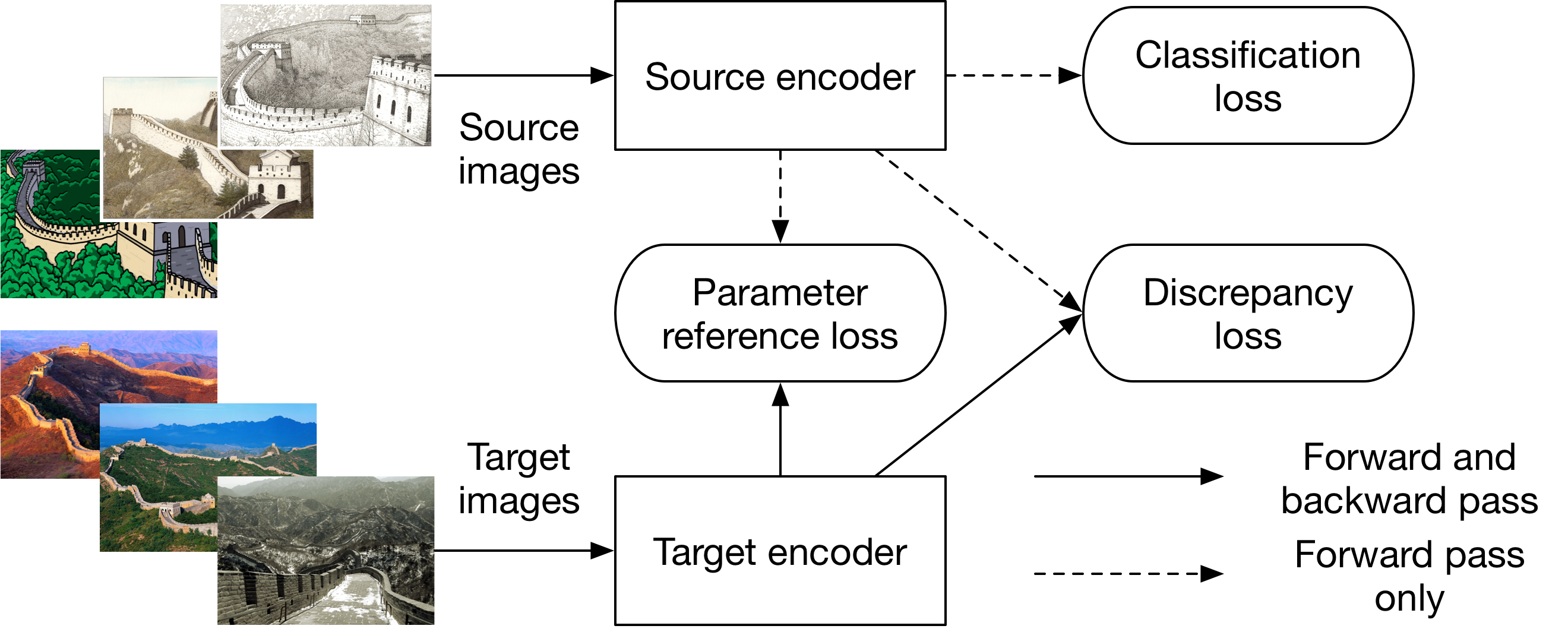}
\end{center}
\vspace{-0.3cm}
   \caption{The diagram of the adaptation phase for the proposed method. A parameter reference loss is computed between the parameters of the source domain encoder and those of the target domain encoder. The dotted line means only the forward pass is needed. For the naive PRL version, learning is disabled in the source domain encoder during the complete adaptation phase, while for other variants of PRL the source domain encoder may also be trained using classification loss, discrepancy loss, and the parameter reference loss. For a simple and fair comparison, the classifier model itself is always fixed during the adaptation.}
\label{fig:prl}
\vspace{-0.4cm}
\end{figure}

Figure~\ref{fig:prl} shows the diagram of the proposed method. We add an extra loss term, the parameter reference loss $\mathcal{L}_{PR}$, on the target domain model as a regularizer. PRL is defined as the $L_1$ loss between the parameters of the source domain encoder denoted as $P_S$, and those of the target domain model denoted as $P_T$. We call it ``reference loss'' because we treat the parameters of the source domain as a reference, which will be utilized during the training instead of only used for initialization. 

The intuition for designing this loss term is three fold: 1) we want to build a connection between the label classifier and the target domain encoder, while there is no direct connection between these two components available; 2) we want to selectively transfer the knowledge learned in the source domain through the parameters, instead of reusing all of the learned parameters like weight sharing. 3) as a result of the constraint from the reference loss, the training is expected to be more stable than independent source and target domain encoders. 

The formal definition of the PRL is as follows:
\begin{align}
	\mathcal{L}_{PR}(E_S, E_T) = \sum_{i=1}^{N_{P}}{\|p_{T_i} - p_{S_i}\|_1},
\end{align}%
where $N_{P}$ denotes the number of parameters in $E_S$ and $E_T$, while $p_{T_i} \in P_T$ and $p_{S_i} \in P_S$ are corresponding parameters of the target encoder and source encoder.

The reason to choose $L_1$ loss instead of $L_2$ loss is that the property of $L_1$ loss makes the connection between the two domain models sparse, which can be seen as selection of keeping the parameters. These connections allow the discriminative ability learned from the label classifier to transfer to the target domain features. On the other hand, the $L_1$ loss also allows for relatively large variations in the other parameters. In this sense, it still has enough flexibility for the model to learn domain invariant features.

Combined with the previously defined MMD loss, the objective function to optimize for the target domain encoder is:
\begin{align}
	\mathcal{L}_{T_{enc}} = \mathcal{L}_{MMD} + \mathcal{L}_{PR}.
\end{align}

\subsection{Variants of PRL}
In the \textbf{naive PRL} setting, we add a new loss term to the target domain encoder. Since the reference parameters (parameters of the source domain encoder) are fixed during the adaptation, as the training continues, the MMD loss is decreasing and the relative weight of the parameter reference loss is increased. As a result, during the later phase of adaptation training, the PRL plays the leading role and the influence from the MMD loss becomes smaller. On one hand, such a property makes the training quite stable; on the other hand, it also tends to prevent the MMD loss (representing the domain discrepancy) from further decreasing in the later training phase. To solve this problem, we further propose several variants of PRL in the remaining part of this section.

\textbf{Simultaneous PRL} This variant of PRL enables the learning of the source domain encoder as well during the adaptation phase. Hence the parameter changes are more flexible than the Naive PRL. The objective function for the source domain encoder is composed of classification loss, MMD loss and PRL:
\begin{align}
	\mathcal{L}_{S_{enc}} = \mathcal{L}_{CLS} + \mathcal{L}_{MMD} + \mathcal{L}_{PR}.
\end{align}

\textbf{Warm-up PRL} This variant of PRL disables the learning of the source domain encoder at the beginning of adaptation. After the MMD loss is decreased to a small value and stops further descreasing, the learning of the source domain encoder is enabled again, until the end of the adaptation phase. This modification from the Simultaneous PRL is intended to prevent the unstable training behavior during the early phase of adaptation.

\textbf{In-turn PRL} This variant of PRL repeatedly disables the learning of the source domain encoder for $k$ epochs, and then enables the learning for another $k$ epochs. The intuition of this strategy is that disabling the learning updates on the source encoder means that the target domain encoder can have a reference constraint to avoid unstable behavior and unintentional dramatic change, while enabling the learning updates on the source encoder means that the source domain model can also use the current target domain model as a reference, allowing a progressive but gradual change of the parameters for both domains. 

\section{Experimentation}
\label{experiment}
In this section, we first describe the domain adaptation datasets that we use for evaluation, and then we compare the different variants of PRL and baseline models on these datasets.

\subsection{Datasets}
To sufficiently evaluate the proposed method with other state-of-the-art methods, we adopt the widely used DA dataset Office~\cite{gong2012geodesic}. To further evaluate the method on more challenging domains, we also try the method on a relatively new dataset: LandmarkDA~\cite{csurka2017discrepancy}. The details of these datasets are described below.

\textbf{Office-31}~This is a classic domain adaptation dataset with three different domains: Amazon, DSLR and Webcam, with 31 classes for each domain. Among the three domains, DSLR and Webcam have a very similar data distribution, thus adaptation on these two domains is easier than the other combinations. We evaluate the overall accuracy for all available configurations (6-direction domain adaptation).

\textbf{LandmarkDA}~This is a very new dataset for visual domain adaptation. It also includes three different domains, photos, paintings and drawings with 25 classes in total. The differences between these domains are much larger than the above datasets, hence it is useful to evaluate the adaptation method in domains with more diversity.

\subsection{Experimental setup}
For all of the datasets, we construct the domain adaptation task with one source domain and one target domain for every possible combination. The performance of the model is evaluated by the overall classification accuracy on the target domain. The Office-31 and LandmarkDA datasets share the same experimental setting including the base model architecture.

There are many design options for an unsupervised domain adaptation model. On one hand, it provides us more probability to improve the model; on the other hand, it makes the comparison of different methods more complicated. To make a fair comparison of different methods, we insist on using the same design options for all the baselines and proposed methods except for the key feature of the methods. The following settings will be used for all methods compared in the experiments.

\textbf{Base architecture}
The study of Csurka \etal~\cite{csurka2017discrepancy} clearly shows that different deep neural networks (DNNs) have large performance variations even when using the same UDA methods. Since the focus of our work is not to improve the architecture of the DNNs, we select the very basic and most widely used AlexNet~\cite{krizhevsky2012imagenet} as the base model for all experiments. Similar to many existing works, we also adopt the AlexNet model pre-trained on ImageNet~\cite{russakovsky2015imagenet} to accelerate the initial supervised learning phase on the source domain.
We are aware that freezing certain layers when fine-tuning the pre-trained model on the new tasks/domains might help improve the performance on some datasets, however, as there are many options for selecting which layers to freeze, we choose to avoid this extra variance by simply fine-tuning all of the parameters of the base model. 
In fact, the idea of freezing layers does not contradict the PRL, since we can still easily apply the PRL on the layers that are not frozen. 

\textbf{Implementation details}
To use the pre-trained AlexNet model on the domain adaptation datasets, we replace the final fully-connected (FC) layer of the original model with a new randomly initialized FC layer suitable for the number of classes on the datasets, \eg, a FC layer with 31-dimensional outputs for the Office-31 dataset. The first baseline model is simply fine-tuning the pre-trained model on the source domain, and we name the encoder part of this model $M_{source}$. All other baselines and variants of PRL are based on this model, and the classifier part of this model is fixed for all adaptation procedures.

The other two baselines are $M_{single}$, which uses a single encoder for both domains, and $M_{double}$, which uses independent encoders for the source and target domains respectively. All these encoders are initialized with the parameters of $M_{source}$. These baselines are actually the unified versions of existing methods like DDC~\cite{tzeng2014deep} and ADDA~\cite{tzeng2017adversarial}, using the same settings for simple comparison.

\textbf{Hyper-parameters}
The Gaussian kernel width for the MMD loss is set to 50000 for all methods and domains. This value was obtained by grid search without accessing the target domain labels. Note that we only need to find a value such that the MMD loss continuously decreases (in an unsupervised manner). We used the fixed optimizer with a learning rate of 0.0001 and a weight decay of 0.00002. These particular values can be chosen based on the performance on the source domain, where labels are available. We used a mini-batch size of 256 for training the $M_{source}$ and a size of 128 for adaptation.

\textbf{PRL variants}
The main difference of PRL variants from the baseline models is the parameter reference loss term. There is one hyper-parameter related to this loss, namely the reference weight. However, this hyper-parameter can also be easily selected by observing the change of MMD loss and the PRL. Using a larger value at the beginning usually does not harm the performance, since it prevents the parameters to change significantly from the original model. Then if we observe that the MMD loss is not decreasing, we can choose a smaller reference weight to allow the discrepancy loss to be minimized. In our experiments, the reference weight is set to 10 and 100 for the Office dataset and LandmarkDA dataset respectively.

\begin{table}[t]
\centering
\begin{tabular}{@{\extracolsep{-0.5pt}}lccccccc}
\toprule
             & A2D  & A2W  & W2A  & W2D  & D2A  & D2W  \\
\midrule
$M_{source}$ & 57.2 & 51.7 & 36.3 & 97.4 & 37.8 & 92.8 \\
$M_{single}$ & 53.6 & 54.0 & 40.0 & 97.1 & 41.3 & 93.1 \\
$M_{double}$ & 49.8 & 35.3 & 31.6 & 87.1 & 36.5 & 89.6 \\
\midrule
PRL($L_2$)      & 62.2 & 46.3 & 33.3 & 92.2 & 38.3 & 91.1 \\
PRL          & \bfseries 64.5 & 58.1 & 39.5 & 96.0 & 39.3 & 92.0 \\
\midrule
simul        & 57.2 & 57.1 & 39.7 & \bfseries 98.6 & 39.1 & \bfseries 93.8 \\
warm-up      & \bfseries 64.5 & 60.0 & 40.0 & 98.4 & 40.6 & 93.7 \\
in-turn      & 63.9 & \bfseries 61.6 & \bfseries 40.4 & \bfseries 98.6 & \bfseries 41.6 & 93.6 \\
\bottomrule
\end{tabular}
\caption{Evaluation on Office-31 using the model after training 50 epochs (latest snapshot of the model), \ie, without model selection. A: Amazon, D: DSLR, W: Webcam.}
\label{tab:reasonable}
\end{table}   

\begin{table}[t]
\centering
\begin{tabular}{@{\extracolsep{-0.5pt}}lccccccc}
\toprule
             & A2D  & A2W  & W2A  & W2D  & D2A  & D2W  \\
\midrule
$M_{source}$ & 57.2 & 51.7 & 36.3 & 97.4 & 37.8 & 92.8 \\
$M_{single}$ & 58.8 & 57.5 & 40.2 & 98.6 & 41.5 & \bfseries 95.2 \\
$M_{double}$ & 56.4 & 55.2 & 35.1 & 96.2 & 39.2 & 90.8 \\
\midrule
PRL($L_2$)      & 62.7 & 56.6 & 36.7 & 96.2 & 39.6 & 91.2 \\
PRL          & \bfseries 64.9 & 61.8 & 40.4 & 98.0 & 40.8 & 92.7 \\
\midrule
simul        & 63.3 & 58.6 & 40.3 & \bfseries 98.8 & 39.3 & 93.8 \\
warm-up      & 64.5 & 62.6 & 40.8 & 98.8 & 41.0 & 94.1 \\
in-turn      & 64.1 & \bfseries 63.1 & \bfseries 41.1 & 98.6 & \bfseries 41.7 & 93.6 \\
\bottomrule
\end{tabular}
\caption{Evaluation on Office-31 using the target domain labels for model selection.}
\label{tab:contradicted}
\vspace{-0.4cm}
\end{table}

\begin{figure}[t]
\begin{center}
   \includegraphics[width=\linewidth]{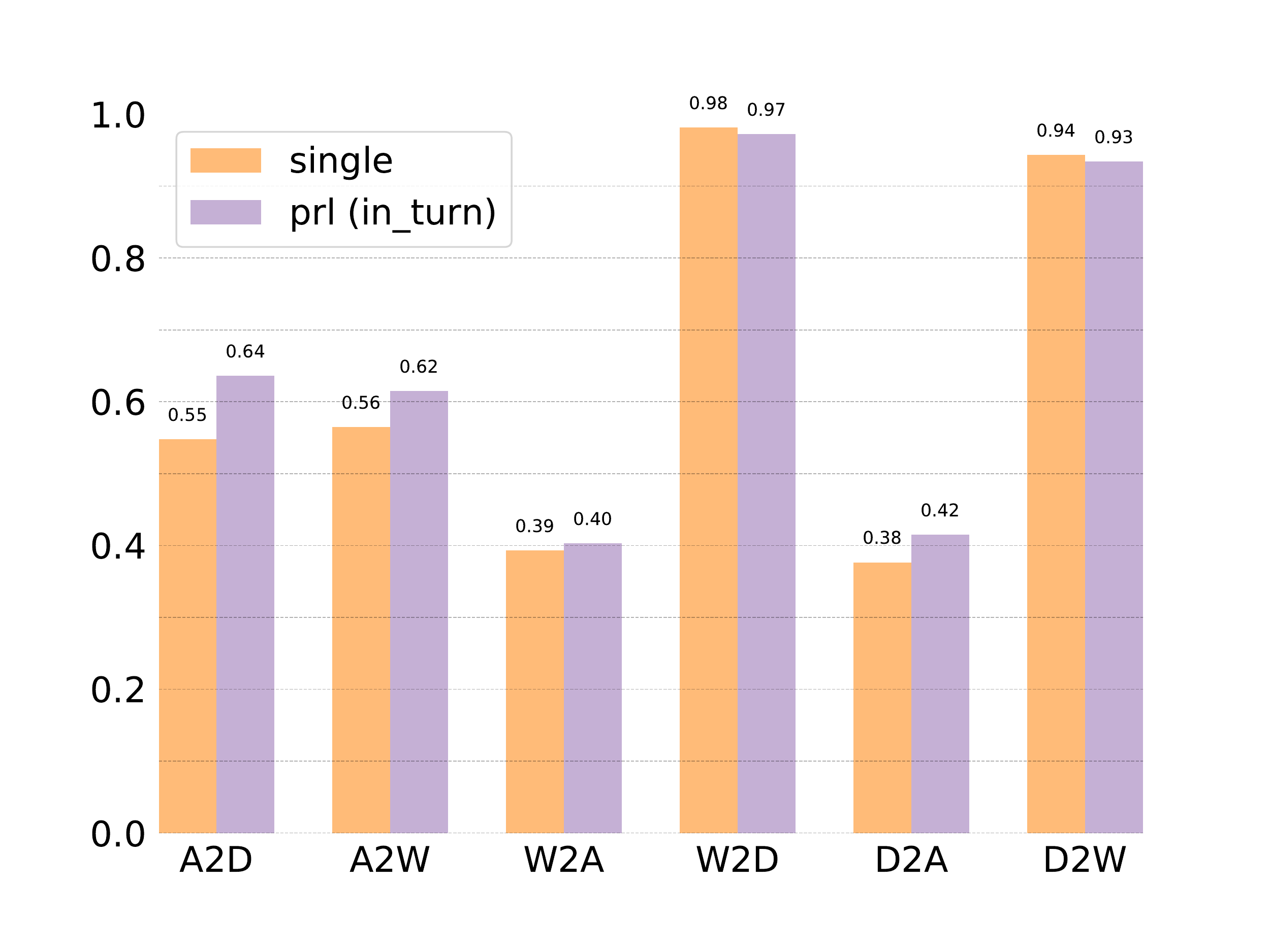}
\end{center}
\vspace{-0.8cm}
   \caption{Accuracy of different models when having the same MMD loss of $0.002$, which is the minimal common value among all experiments.}
\label{fig:discriminative}
\vspace{-0.4cm}
\end{figure}

\begin{figure*}
    \centering
    \begin{subfigure}[b]{0.48\textwidth}
        \centering
        \includegraphics[width=\linewidth]{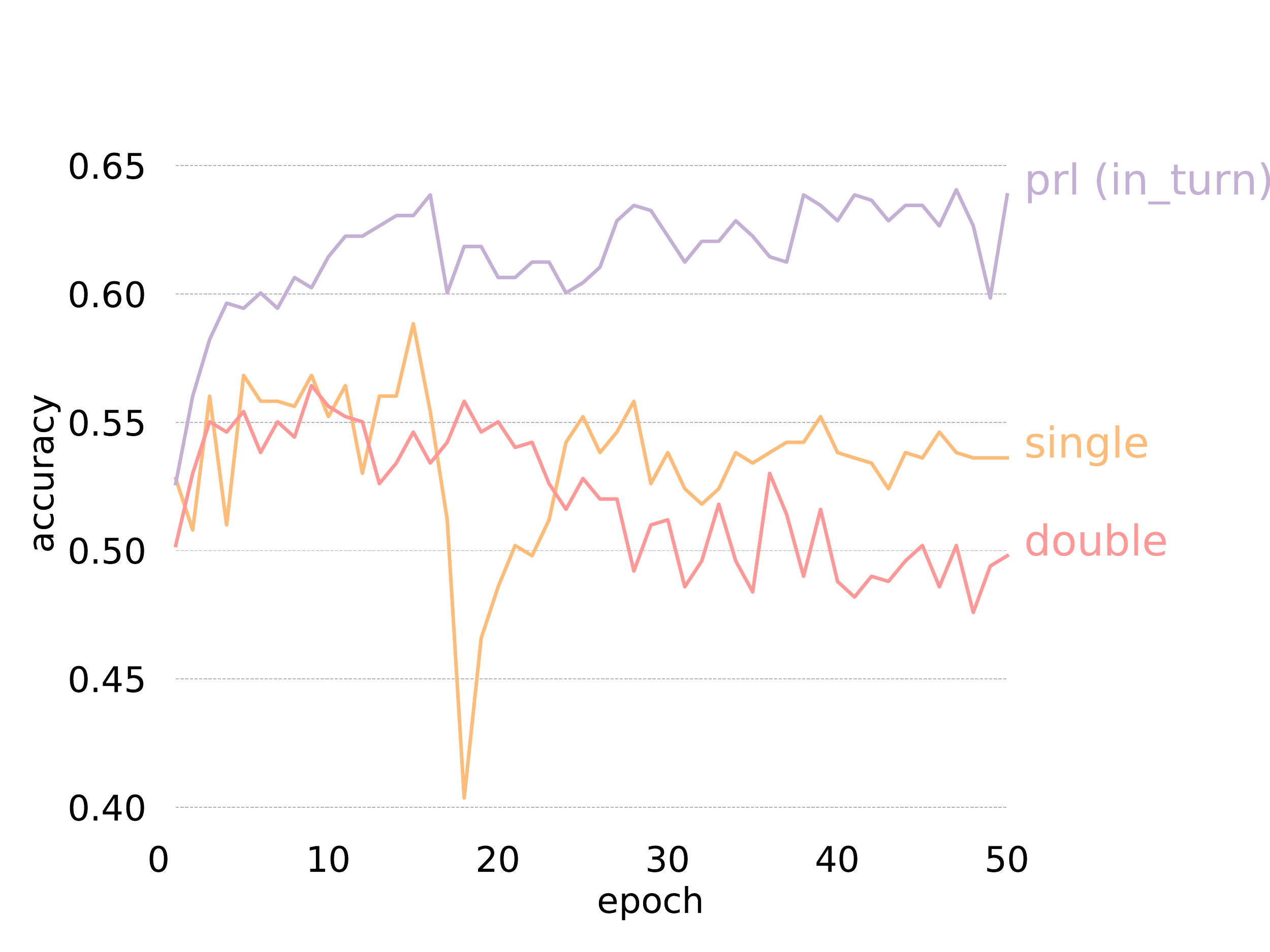} 
        \caption{Amazon to DSLR} \label{fig:a2d}
    \end{subfigure}
    \hfill
    \begin{subfigure}[b]{0.48\textwidth}
        \centering
        \includegraphics[width=\linewidth]{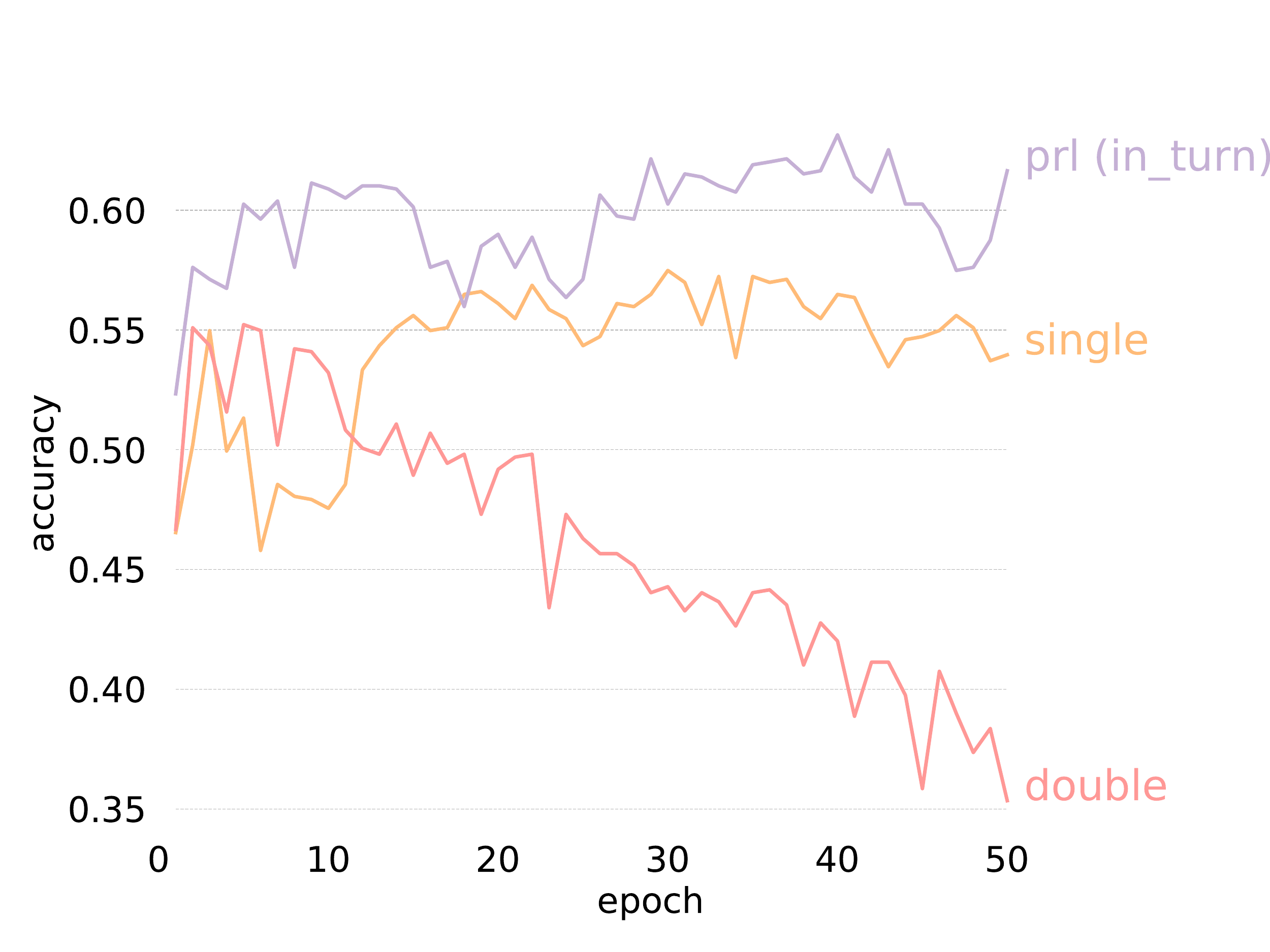} 
        \caption{Amazon to Webcam} \label{fig:a2w}
    \end{subfigure}
    \caption{Accuracy on the target domain during the adaptation training phase. }
\label{fig:stability}
\end{figure*}

\subsection{Analysis}
Table~\ref{tab:reasonable} shows the evaluation results on the Office-31 dataset using the proposed evaluation procedure, \ie, without using the target domain labels for hyper-parameter tuning and model selection.

\textbf{Baseline models}
 $M_{single}$ almost performs better than $M_{source}$ on all adaptation directions, while $M_{double}$ performs worse than the $M_{source}$ on all adaptation configurations. The results indicate that too much flexibility in the double stream encoders can even harm the performance.

\textbf{$L_1$ versus $L_2$}
The results show that $L_1$ loss performs much better than the $L_2$ loss. Though more evidence is needed, these results suggest that selectively fixing some parameters while allowing other parameters to vary relatively largely might be better than allowing many arbitrary changes on all parameters. 

\textbf{Variants of PRL}
The simultaneous PRL does not work so well in general due to the large flexibility during the early training phase, though it is still generally better than $M_{single}$. On domains that are already highly similar before the adaptation (DSLR and Webcam), the performance of simultaneous PRL is good, but the small differences are not enough to prove the superiority over other methods. Warm-up and in-turn PRL perform better than all other candidates, of which in-turn PRL is more stable and has slightly better performance on the Office dataset. Figure~\ref{fig:stability} demonstrates the superior stability of PRL during the adaptation training phase.
When comparing the variants of PRL to $M_{single}$, we have observed that when the source domain has a relatively large scale data (A2D and A2W), the performance of PRL is significantly better than that of $M_{single}$ and other baselines, which is evidence that PRL can utilize the knowledge learned from the source domain more effectively. 

\begin{table*}[t]
\centering
\begin{tabular}{@{\extracolsep{7.35pt}}lcccccccc}
\toprule
             & Ph $\rightarrow$ Pa   & Ph $\rightarrow$ Dr  & Pa $\rightarrow$ Ph  & Pa $\rightarrow$ Dr  & Dr $\rightarrow$ Ph  & Dr $\rightarrow$ Pa  & Avg\\
\midrule
$M_{source}$ & 66.5 & 55.7 & 79.5 & 63.7 & 74.3 & 61.0 & 66.8\\
$M_{single}$ & 69.9 (70.3) &  64.4 (65.2) & 81.1 (82.5) & \bfseries  70.9 (71.5) & 74.1 (76.0) & 67.5 (68.6) & 71.3 (72.3) \\
$M_{double}$ & 48.7 (54.1) & 35.6 (43.3) & 60.1 (74.5) & 42.6 (52.2) & 43.4 (59.2) & 36.4 (54.8) & 44.5 (56.4) \\
\midrule
warm-up      & \bfseries 70.1 (70.5) & \bfseries 65.6 (66.0) & 81.8 (82.4) & 70.6 (70.8) & \bfseries 76.2 (77.5) & \bfseries 68.3 (68.9) & \bfseries 72.1 (72.7) \\
in-turn      & 69.6 (70.0) & 63.9 (64.5) & \bfseries 82.2 (82.5) & 69.2 (69.7) & 75.6 (77.4) & 67.8 (68.9) & 71.4 (72.2) \\
\bottomrule
\end{tabular}
\captionsetup{width=\textwidth}
\vspace{-0.2cm}
\caption{Evaluation on LandmarkDA using different evaluation principles. Two results are reported for each model except for $M_{source}$. The values on the left are obtained using the latest epoch ($100$ epochs in total) of each model. The values in the brackets are obtained using target domain labels, \ie, the best result during training. Ph: Photo, Pa: Painting, Dr: Drawing.}
\label{tab:landmark}
\vspace{-0.4cm}
\end{table*}
\textbf{Discriminative ability}
As shown in Figure~\ref{fig:discriminative}, when $M_{single}$ and in-turn PRL have a very similar and small MMD loss, there is a large margin of performance gap between these two models on A2D and A2W adaptation tasks. These results support our claim that existing methods lack an adequate way to preserve the discriminative ability of the target domain features. In contrast, PRL is able to generate better discriminative features by making use of learned parameters more efficiently. The results also show that the performance of $M_{single}$ (parameter sharing) depends on the similarity between the source domain and the target domain (the performance is better on W2D and D2W tasks), however, fully sharing the parameters may not be a good idea because not all of the knowledge learned from the source domain is applicable to the target domain. The advantage of using $L_1$ loss to achieve selective knowledge transfer results in the superior performance of PRL on domains with more diversity.

\textbf{Influence of evaluation procedures} Now that we have seen the results obtained using our proposed evaluation procedure, it is interesting to see how the results compare to those obtained by traditional evaluation procedures. Table~\ref{tab:contradicted} lists the evaluation results obtained by using the target domain labels for model selection. We see that the baseline methods are significantly influenced by the evaluation methods. In contrast, we observe that the PRL has more stable performance during training, and hence is expected to be more suitable to real world applications in which no target domain labels are available.

\textbf{Results on LandmarkDA}
The PRL variants have slightly better performance and stability compared to $M_{single}$, and is significantly better than $M_{double}$ on the LandmarkDA dataset, as shown in Table~\ref{tab:landmark}. The results are generally consistent with those on the Office dataset. Besides, even the averaged accuracy ($72.1$) from the latest epoch (without model selection) achieves superior performance to state-of-the-art methods ($69.1$), as reported in~\cite{csurka2017discrepancy}, which also uses AlexNet to extract deep features and MMD to minimize domain discrepancy.
\vspace{-0.2cm}
\section{Conclusion}
We observe that existing Siamese network-based domain adaptation methods have limited ability to produce target domain features that are able to retain discriminative ability. We then propose a novel parameter reference loss that encourages the parameters of the target domain encoder to partially remain close to those of the source domain encoder, which has a direct connection to the classifier. This allows a more flexible and selective use of parameters learned from the source domain during the domain adaptation phase compared to full parameter sharing.
Our experiments show that even the naive approach of using the pre-trained encoder parameters as a fixed reference during domain adaptation can improve the adaptation performance as well as the training stability. These results are in agreement with our hypothesis that only using the learned encoder parameters for initializing and fine-tuning the target encoder on a new domain/task is an inefficient use of resources. 

In addition, we argue that existing UDA evaluation procedures can be contradictory to the requirements that such models are expected to meet in real-world usage. We therefore propose to use a simple but more reasonable evaluation procedure for hyper-parameter tuning and model selection without access to target domain labels. We argue that the key requirement is to make the UDA training more stable, which is actually more important in real applications. Our experimental results indicate that even in such a strict situation, our proposed method still manages to achieve a stable and superior performance than the other existing baselines.

{\small
\bibliographystyle{ieee}
\bibliography{egbib}
}

\appendix

\section{Apply PRL to existing methods}
\begin{table*}[!htbp]
\centering
\begin{tabular}{@{\extracolsep{4.5pt}}lcccccccc}
\toprule
             & A $\rightarrow$ D   & A $\rightarrow$ W  & W $\rightarrow$ A  & W $\rightarrow$ D  & D $\rightarrow$ A  & D $\rightarrow$ W  & Avg\\
\midrule
CNN & $63.8 \pm 0.5$ & $61.6 \pm 0.5$ & $49.8 \pm 0.4$ & $99.0 \pm 0.2$ & $51.1 \pm 0.6$ & $95.4 \pm 0.3$ & $70.1$\\
DDC & $64.4 \pm 0.3$ & $61.8 \pm 0.4$ & $52.2 \pm 0.4$ & $98.5 \pm 0.4$ & $52.1 \pm 0.8$ & $95.0 \pm 0.5$ & $70.6$\\
DAN & $65.8 \pm 0.4$ & $63.8 \pm 0.4$ & $51.9 \pm 0.5$ & $98.8 \pm 0.6$ & $52.8 \pm 0.4$ & $94.6 \pm 0.5$ & $71.3$\\
Deep CORAL & $66.8 \pm 0.6$ & $66.4 \pm 0.4$ & $51.5 \pm 0.3$ & $99.2 \pm 0.1$ & $52.8 \pm 0.2$ & $95.7 \pm 0.3$ & $72.1$\\
\midrule
CNN (ours) & $61.8 \pm 0.6$ & $58.7 \pm 0.6$ & $46.6 \pm 0.9$ & $98.9 \pm 0.4$ & $47.2 \pm 0.4$ & $93.8 \pm 0.3$ & $67.8$\\
single (latest) & $57.7 \pm 1.6$ & $57.3 \pm 1.2$ & $46.8 \pm 0.7$ & $98.9 \pm 0.4$ & $46.4 \pm 0.4$ & $95.7 \pm 0.2$ & $67.1$\\
in-turn (latest) & $64.6 \pm 0.5$ & $\bm{65.8} \pm 0.8$ & $47.8 \pm 0.7$ & $99.1 \pm 0.2$ & $48.1 \pm 0.7$ & $95.3 \pm 0.5$ & $70.1$\\
warm-up (latest) & $\bm{65.7} \pm 1.6$ & $64.7 \pm 0.5$ & $\bm{49.1} \pm 0.5$ & $\bm{99.2} \pm 0.2$ & $\bm{49.6} \pm 0.8$ & $\bm{96.0} \pm 0.1$ & $\bm{70.7}$\\
\midrule
single (best) & $65.5 \pm 0.2$ & $61.8 \pm 1.0$ & $47.1 \pm 0.7$ & $99.2 \pm 0.3$ & $46.7 \pm 0.5$ & $96.1 \pm 0.1$ & $69.4$\\
in-turn (best) & $65.7 \pm 1.1$ & $\bm{67.2} \pm 0.9$ & $48.6 \pm 0.8$ & $\bm{99.5} \pm 0.1$ & $48.9 \pm 0.7$ & $96.0 \pm 0.2$ & $71.0$\\
warm-up (best) & $\bm{67.2} \pm 0.5$ & $65.7 \pm 0.6$ & $\bm{49.3} \pm 0.4$ & $99.4 \pm 0.1$ & $\bm{50.6} \pm 1.1$ & $\bm{96.2} \pm 0.2$ & $\bm{71.4}$\\

\bottomrule
\end{tabular}
\captionsetup{width=\textwidth}
\caption{Target domain accuracy (mean $\pm$ std) on the Office dataset, comparing state-of-the-art methods with PRL using the same architecture as of Deep Domain Confusion (DDC)~\cite{tzeng2014deep}. The first group of results (top 4 rows) are directly borrowed from Deep CORAL~\cite{sun2016deep}. Note that even the results of the same methods are reported slightly different in existing work, thus we use them just for reference. Besides, the performance of Deep Adaptation Networks (DAN)~\cite{long2015learning} is obtained by using multi-kernel MMD on a single layer for direct comparison, as stated in ~\cite{sun2016deep}. CNN (ours) is our implementation of the base model that is only trained on the source domain, without adaptation. The second group of results (middle 4 rows) report the performance of latest snapshots of the models; while the third group of results (bottom 3 rows) list the performance of the best snapshots for the models. We highlight the best results for the latest models and best models separately.}
\label{tab:sota}
\end{table*}

We have been focusing on comparing Parameter Reference Loss (PRL) with different baselines (weight sharing or independent encoders) under the same base model architecture and design options in the main paper. Here we instead show an example of applying PRL to existing unsupervised domain adaptation (UDA) methods, and compare the performance with several state-of-the-art methods based on the discrepancy loss.

We select one of the most basic UDA method based on deep neural networks, Deep Domain Confusion (DDC)~\cite{tzeng2014deep}, to be combined with PRL. Different from the model used in the main paper, we add an extra adaptation layer as a bottleneck layer after the $fc7$ layer of AlexNet~\cite{krizhevsky2012imagenet}. The adaptation layer is a fully-connected layer with 256-dimension outputs, following~\cite{tzeng2014deep}. The only difference between our model and DDC is the use of PRL instead of weight sharing between the source and target domain encoders.

To facilitate direct comparisons between our models and the existing methods, we tried to improve the model trained only on the source domain. By using techniques like freezing layers during the source domain training phase, we managed to obtain a model much better than the one used for comparing different baselines and PRL variants, however, its performance is still weaker than the one reported in previous work. 

For all experiments of PRL variants, the training procedure is exactly the same as described in the main paper. Besides, we have modified some hyper-parameters according to the change of features caused by the extra adaptation layer. Specifically, we use the kernel width 1000 for MMD, and reference weight 100 for PRL in the following experiments. All these hyper-parameters are selected without access to the target domain labels. 

\subsection{Analysis}
Table~\ref{tab:sota} shows the experimental results on the Office dataset. The first four columns are from Deep CORAL~\cite{sun2016deep}. Note that the implementation of our base model performs worse than the one used in previous work ($67.8$ vs $70.1$). The middle columns of results are obtained by selecting the latest snapshot of the model, while the bottom columns of results are from the best snapshot during the training phase, \ie, using model selection with the target domain labels.

\textbf{Latest models} We first compare the latest models of different methods. The results are consistent with the previous experiments: in-turn and warm-up PRL perform better than single-encoder model (DDC), especially on the A $\rightarrow$ D and A $\rightarrow$ W tasks.

\textbf{Comparison with existing methods} We can clearly see that warm-up PRL outperforms DDC even with a weaker CNN base model ($70.7$ vs $70.6$). Considering the difference of the CNN base model, the improvements in averaged accuracy is $2.9$ ($70.7 - 67.8$) vs $0.5$ ($70.6 - 70.1$). Again, the advantages of PRL is especially strong on the A $\rightarrow$ D and A $\rightarrow$ W tasks, where PRL also outperforms DAN (single layer multi-kernel version) and is comparable to Deep CORAL. Such results may support the hypothesis that PRL can take advantage of knowledge learned from the source domain more efficiently, particularly when the source domain has a larger amount of data. (Number of images, Amazon: 2817, DSLR: 498, Webcam: 795).

\textbf{Best models} Obviously the best models have better performance compared to the latest models. What is important is to see the differences caused by the model selection method (with or without access to the target domain labels). We found that PRL variants are in general more stable than the single-encoder model (DDC) regarding to the model selection changes. These models also achieve state-of-the-art performance on several tasks, however, we consider it not proper to be used for evaluating the method. Hence we report the results here only for reference. 

\subsection{A general technique for UDA}
The main objective of conducting above experiments is to show the potential of PRL to improve existing UDA methods. Due to the simplicity of PRL, it can be easily applied to almost all Siamese network-based models (\eg, DAN, Deep CORAL, \etc) with minor effort on modifying existing code. In fact, PRL can be used as a basic technique like weight sharing, to provide another kind of flexible regularization. We are also interested in utilizing PRL in tasks other than domain adaptation, and would like to investigate more on efficient use of learned parameters.

\end{document}